\documentclass{article}
\usepackage{microtype}
\usepackage{graphicx}
\usepackage{subcaption}
\usepackage{booktabs} 
\usepackage{hyperref}

\usepackage[preprint]{icml2026}

\usepackage{amsmath}
\usepackage{amssymb}
\usepackage{mathtools}
\usepackage{amsthm}
\usepackage[capitalize,noabbrev]{cleveref}
\theoremstyle{plain}

\theoremstyle{definition}

\theoremstyle{remark}

\usepackage[textsize=tiny]{todonotes}
\usepackage[table]{xcolor}
\usepackage{xspace}
\newcommand{\themodel}{NAG\xspace}
\icmltitlerunning{NAG: A Unified Native Architecture for Encoder-free Text-Graph Modeling in Language Models}

\begin{document}

\twocolumn[
  \icmltitle{NAG: A Unified Native Architecture for Encoder-free Text-Graph Modeling in Language Models}
  \icmlsetsymbol{equal}{*}
  \begin{icmlauthorlist}
    \icmlauthor{Haisong Gong}{equal,zdh,rgzn}
    \icmlauthor{Zhibo Liu}{equal,jcxy,zdh}
    \icmlauthor{Qiang Liu}{zdh,rgzn}
    \icmlauthor{Shu Wu}{zdh,rgzn}
    \icmlauthor{Liang Wang}{zdh,rgzn}
  \end{icmlauthorlist}
  \icmlaffiliation{zdh}{New Laboratory of Pattern Recognition (NLPR), Institute of Automation, Chinese Academy of Sciences, Beijing, China}
  \icmlaffiliation{rgzn}{School of Artificial Intelligence, University of Chinese Academy of Sciences (UCAS), Beijing, China}
  \icmlaffiliation{jcxy}{School of Advanced Interdisciplinary Sciences, UCAS, Beijing, China}

  % \icmlcorrespondingauthor{Shu Wu}{shu.wu@nlpr.ia.ac.cn}
  
  % \icmlcorrespondingauthor{Firstname2 Lastname2}{first2.last2@www.uk}
  % You may provide any keywords that you find helpful for describing your
  % paper; these are used to populate the "keywords" metadata in the PDF but
  % will not be shown in the document
  \icmlkeywords{Graph, Language Models}
  \vskip 0.3in
]

% this must go after the closing bracket ] following \twocolumn[ ...

% This command actually creates the footnote in the first column listing the
% affiliations and the copyright notice. The command takes one argument, which
% is text to display at the start of the footnote. The \icmlEqualContribution
% command is standard text for equal contribution. Remove it (just {}) if you
% do not need this facility.

% Use ONE of the following lines. DO NOT remove the command.
% If you have no special notice, KEEP empty braces:
% \printAffiliationsAndNotice{}  % no special notice (required even if empty)
% Or, if applicable, use the standard equal contribution text:
\printAffiliationsAndNotice{\icmlEqualContribution}

\begin{abstract}
Prevailing methods for integrating graphs into Language Models (LMs) typically rely on a segregated architecture: external Graph Neural Networks (GNNs) encode structural topology, while LMs process textual semantics. We argue this approach is suboptimal for text-graphs: it creates a conceptually disjointed interaction paradigm. By segregating structural encoding from semantic processing, these systems must perform a complex implicit alignment between abstract graph tokens and concrete textual elements. Challenging the necessity of external encoders, we propose NAG (Native Architecture for Graphs), a unified framework that internalizes graph processing within the LM's native manifold. Instead of bridging disparate embedding spaces, NAG repurposes the self-attention mechanism to enforce topological dependencies and recalibrates positional IDs to ensure structural equivalence. This allows the model to harness its intrinsic linguistic capability to simultaneously comprehend node and edge content alongside structural topology. We introduce two efficient implementations: NAG-Zero for absolute preservation of the base model’s linguistic capabilities, and NAG-LoRA for enhanced structural adaptation. Experiments across diverse graph tasks validate that NAG achieves robust graph comprehension without the overhead of external encoders, offering a simpler, more coherent paradigm for text-graph modeling.
\end{abstract}

%================================================================
\section{Introduction}

\begin{figure}[t]
  % \vskip 0.2in
  \begin{center}
    \centerline{\includegraphics[width=\columnwidth]{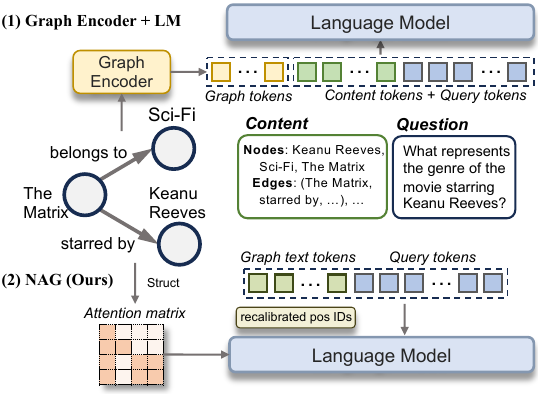}}
    \caption{
    Comparison of graph encoding paradigms: (Top) Dual-pathway architecture using an external graph encoder to generate graph tokens, which are prepended to the text sequence for the language model. (Bottom) Our proposed \themodel framework, which eliminates the external encoder and models graph structure natively via a topology-aware attention mechanism and recalibrated positional indexing within the language model.
    }
    \label{fig:introfig}
  \end{center}
  % \vspace{-0.4cm}
\end{figure}

Language Models (LMs) have demonstrated remarkable proficiency in processing sequential data, serving as the foundation for modern natural language understanding. However, real-world knowledge is not always sequential; it frequently manifests as text-graphs where nodes and edges are intrinsically composed of rich textual information (e.g., knowledge graphs, citation networks, and dialogue trees). Empowering LMs to comprehend these non-Euclidean structures is essential for advancing their reasoning capabilities beyond linear text.

A prevailing approach to this challenge parallels architectures found in multi-modal modeling. This paradigm typically adopts a dual-pathway architecture: an external Graph Neural Network (GNN) serves as a dedicated structure encoder, while the LM independently acts as the semantic processor~\cite{he2024g,perozzi2024let}. In practice, the GNN encodes graph inputs, initialized with either simple heuristic features~\cite{liu2024can} or embeddings from auxiliary models like S-BERT~\cite{reimers2019sentence}, into a sequence of latent ``graph tokens.'' These tokens are then projected into the LM’s input space and prepended to the textual content of nodes and edges. Essentially, the LM receives the raw text sequence accompanied by a set of ``soft prompts" intended to represent the graph's topology, as illustrated in Figure \ref{fig:introfig} (Top).

While these methods have advanced the field, we argue that this dual-pathway architecture creates a conceptually disjointed interaction paradigm. By segregating structural encoding from semantic processing, these systems necessitate an implicit alignment between abstract ``graph tokens'' rooted in heterogeneous feature spaces and concrete textual elements. This creates a fundamental representational disparity, as the architecture must reconcile external signals with the intrinsic textual semantics it is natively primed to process, ultimately resulting in a system that lacks the coherence of a unified native architecture.

This prompts a fundamental question: \textit{Since the content of nodes and edges is text, can we discard the external encoder and model the graph natively within the LM?}

In this paper, we answer affirmatively by proposing \themodel (Native Architecture for Graphs), an encoder-free framework (shown in Figure \ref{fig:introfig}, bottom). \themodel treats nodes and edges as a set of textual units and processes them directly. Our core innovation lies in a specialized attention mechanism: it applies a standard causal mask within each textual element for precise semantic encoding, while simultaneously opening attention pathways between connected elements to enable structural information flow. Complementing this, we employ a recalibrated positional indexing strategy. By assigning equivalent positional IDs to structurally parallel elements, we ensure the LM perceives the graph without the sequential bias inherent in serialization order. Intuitively, this design allows the LM to function as a dual-process engine: its layers act as GNN-like message-passing steps for topology while simultaneously performing semantic encoding, achieving a unified parallel processing of structure and content.

We present two efficient implementations of \themodel. \themodel-Zero utilizes inter-layer gated adapters to achieve zero-interference with the base model, preserving 100\% of its original linguistic capabilities. \themodel-LoRA employs intra-layer low-rank adaptation for scenarios requiring deeper structural-semantic fusion.
We validate the feasibility of this design across diverse graph QA tasks, ranging from fundamental topological inquiries (e.g., shortest path, node degree) to representative real-world textual graphs. Our results demonstrate that \themodel's native modeling is not only a rational design choice but a promising and effective alternative for graph comprehension, eliminating the need for external encoders.

\section{Related Works}

\subparagraph{The Dual-pathway: Graph Neural Networks as Prefixes.} A prominent research paradigm conceptualizes graph structures as a specialized modality, typically employing an auxiliary GNN-based module to extract structural inductive biases and generate graph-aware soft prompts as prefix to guide the LMs~\cite{perozzi2024let,liu2024can}. To bridge the semantic gap between ``graph tokens" and textual instructions in prompts, research employs multi-stage instruction tuning~\cite{graphgpt}, contrastive alignment~\cite{higpt}, cross-modality pooling~\cite{GNP}, and specialized graph-to-text translators~\cite{graphllm,graphtranslator,huang2024graphadapter}. Techniques such as subgraph extraction and retrieval have also been introduced to refine the input space, thereby enhancing the performance and robustness of this design~\cite{he2024g,GNP}. Although these methods successfully leverage the structural modeling strengths of GNNs, they inherently suffer from the architectural bifurcation endemic to the two-stage ``encode-then-align" paradigm. This disjointed nature often results in the dilution of the rich structural content captured by the GNNs, ultimately imposing constraints on the performance of graph reasoning tasks.

\subparagraph{Graph Reasoning via Textual Linearization.} Another avenue explores graph reasoning by transforming structures into linear text sequences. This process involves decomposing complex graphs into serial formats ingestible by LMs. Key decomposition strategies include degree-based ranking~\cite{degreebasedranking}, multi-hop traversals~\cite{GraphICLl,llaga,chen2024exploring}, anchor-based encoding~\cite{NT-LLM}, and random walks~\cite{WALklm,musegraph}. Regarding linguistic representation, representative works examine how text encoding functions and prompt designs impact reasoning performance~\cite{NLgraph,fatemi2023talklikegraphencoding}. Recently, emerging agent-based RAG and Chain-of-Thought frameworks~\cite{graphcot,thinkongraph,graphrunner} represent dynamic extensions and hybridizations of these linearization methods.

\subparagraph{Graph Reasoning via Specialized Transformer.} While linearization offers simplicity, research shows that the native attention of LMs struggles to effectively leverage connectivity information in linearized graphs, regardless of the prompting scheme~\cite{attention-perspective}. Consequently, intrinsic paradigms seek to modify the Transformer architecture. Prevalent strategies involve injecting structural priors directly into the attention scoring matrix to create structure-aware operators~\cite{graphfomer,graphbert,distanceencoding}, or employing structural masking to mimic GNN message-passing locality by enforcing sparsity patterns consistent with graph connectivity~\cite{dwivedi2020generalization, yao-etal-2020-heterogeneous}. However, these prior modifications have primarily been explored within encoder-only architectures tailored for discriminative tasks, rather than the autoregressive decoder-only frameworks that power modern generative LMs. This architectural divergence leaves a gap in adapting topological awareness to open-ended text generation. NAG bridges this gap by synergizing topology-aware attention with structural position calibration, achieving endogenous graph reasoning within the generative manifold.

\begin{figure}[t]
  % \vskip 0.2in
  \begin{center}
    \centerline{\includegraphics[width=\columnwidth]{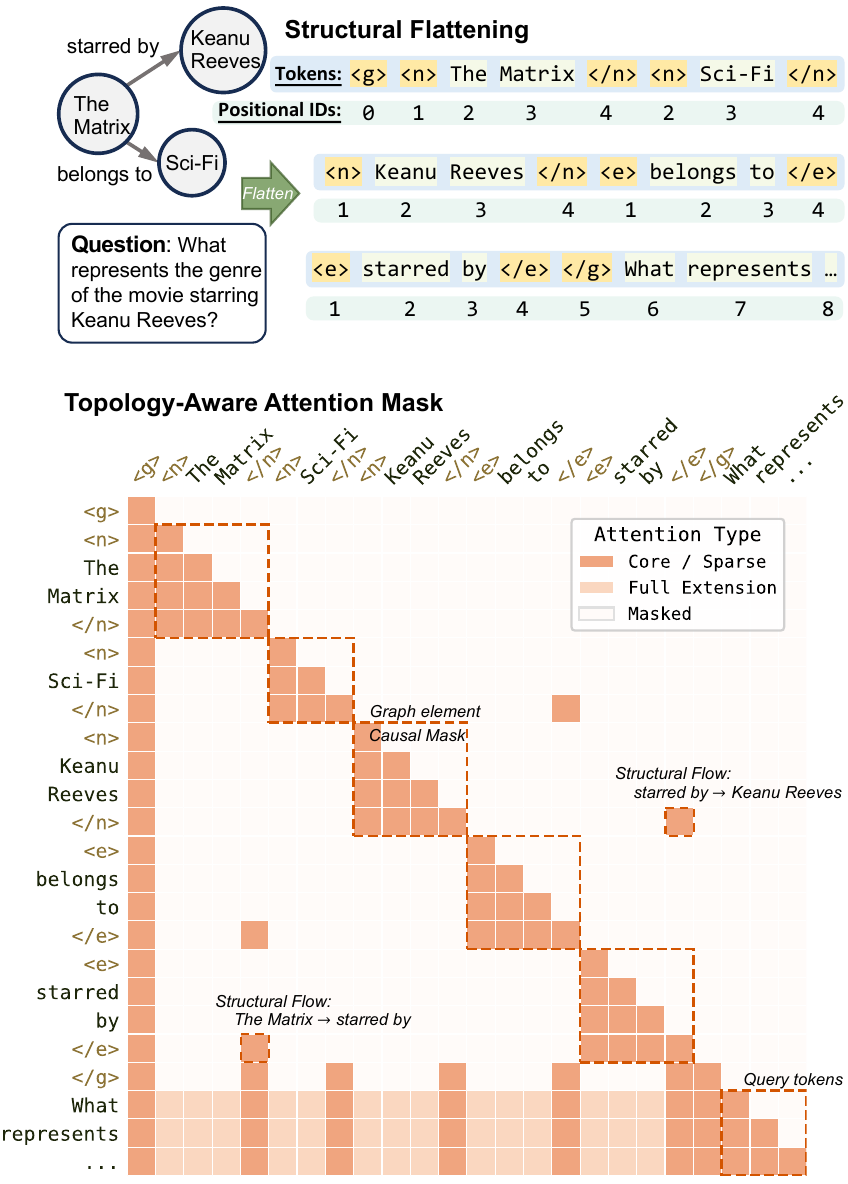}}
    \caption{
Input Construction and Attention Mechanism of \themodel. 
(Top) The graph is flattened with recalibrated positional IDs to ensure structural equivalence. 
(Bottom) The topology-aware mask unifies semantic understanding and structural reasoning: `Core / Sparse' denotes structural dependencies and the sparse query-graph interaction mode, while `Full Extension' represents the additional attention pathways activated in the full interaction mode.
}
    \label{fig:main}
  \end{center}
% \vspace{-0.4cm}  
\end{figure}

\section{The \themodel Framework}

\subsection{Problem Formulation}
\label{sec:problem_formulation}

We define a text-graph as $\mathcal{G} = (\mathcal{V}, \mathcal{E})$, comprising a set of nodes $\mathcal{V}$ and edges $\mathcal{E}$. Each node $v \in \mathcal{V}$ and edge $e \in \mathcal{E}$ is intrinsically composed of a raw text sequence, as shown in Figure~\ref{fig:main}.

To facilitate unified modeling, we establish the \textbf{unified element set} $\mathcal{U} = \mathcal{V} \cup \mathcal{E}$. Within our framework, we explicitly treat edges and nodes as equivalent semantic units 
(graph elements). Consequently, the graph topology is formulated as the connectivity dependencies between these elements.

Given a text-graph $\mathcal{G}$ and a natural language query $Q$, the objective is to generate an answer $A = \{a_1, a_2, \dots, a_{|A|}\}$. We formulate this as a conditional generation task, where the parameterized language model $P_\theta$ models the probability of the target sequence autoregressively:

\begin{equation}
    P_\theta(A \mid \mathcal{G}, Q) = \prod_{t=1}^{|A|} P_\theta(a_t \mid a_{<t}, \mathcal{G}, Q)
\end{equation}

Here, conditioning on $\mathcal{G}$ implies that the generation of each token $a_t$ depends jointly on the semantic content of the unified elements $\mathcal{U}$ and their topological structure.

%=========================================

\subsection{Input Construction via Structural Flattening}
\label{sec:input_construction}

Standard Language Models operate on sequential inputs. To adapt the non-linear structure of $\mathcal{G}$ to this interface, we employ a \textbf{structural flattening} strategy that maps the set of graph elements into a single token sequence $S$.

We introduce three pairs of special tokens to act as semantic boundaries and functional indicators:
\begin{itemize}
    \setlength\itemsep{0em}
    \item \texttt{<n>}, \texttt{</n>}: Wrappers for node content.
    \item \texttt{<e>}, \texttt{</e>}: Wrappers for edge content.
    \item \texttt{<g>}, \texttt{</g>}: Wrappers for the whole graph content.
\end{itemize}

Formally, given the unified element set $\mathcal{U} = \{u_1, u_2, \dots, u_{|\mathcal{U}|}\}$, we serialize them into a sequence bounded by the global tags, with the query $Q$ appended to the end. A crucial premise of our framework is that the physical order of elements in $S$ is \textbf{arbitrary} (e.g., random shuffling or BFS order) and carries no structural implication. The logical connectivity is decoupled from this sequence and is handled exclusively by the topology-aware attention mask (Section \ref{sec:attention}) and recalibrated positional IDs (Section \ref{sec:position}).

\textbf{Illustrative Example.} Consider a graph centered on the movie \textit{The Matrix}, as shown in Figure \ref{fig:main}. The constructed input sequence $S$ is formatted as:
\begin{equation*}
\begin{aligned}
    S = & \left[ \right.  \texttt{<g>}\texttt{<n>The Matrix</n>}\texttt{<n>Sci-Fi</n>} \\
                       & \texttt{<n>Keanu Reeves</n>}\texttt{<e>belongs to</e>} \\
                       & \texttt{<e>starred by</e>}\texttt{</g>} \, ; \, Q \left. \right]
\end{aligned}
\end{equation*}
Note that while we group elements by type in this example for readability, the model is permutation-invariant to the element order. In this layout, the semantic content of each element is encapsulated within its respective tags, preparing them for independent encoding by the LM, while the global structural relationships remain latent until activated by the attention mechanism.

% =====================================

\subsection{Topology-Aware Attention Mechanism}
\label{sec:attention}

While structural flattening organizes the input content, the reconstruction of the logical graph topology is achieved by guiding the LM's encoding behavior. We propose a binary \textbf{Topology-Aware Attention Mask} $M \in \{0, 1\}^{|S| \times |S|}$, where $|S|$ denotes the total number of tokens in the input sequence.

This mask is integrated into the self-attention mechanism of each Transformer block in the LM. Mathematically, setting $M_{i,j}=1$ permits the $i$-th token to attend to (and aggregate information from) the $j$-th token, whereas $M_{i,j}=0$ strictly prevents interaction. As visualized in Figure \ref{fig:main}, we manipulate these visibility constraints to orchestrate information flow across four hierarchical levels. 

For notation, let $\mathcal{T}(u)$ be the set of token indices belonging to element $u \in \mathcal{U}$, and let $\text{hub}(x)$ denote the index of the closing tag for any $x \in \mathcal{U} \cup \{\mathcal{G}\}$ (e.g., \texttt{</n>}, \texttt{</e>}, and \texttt{</g>}).

\textbf{Level 1: Intra-Element Semantic Encoding.}
For tokens within the same element, we apply a standard \textbf{Causal Mask}. Crucially, we enforce strict isolation between distinct elements ($u \neq v$) to ensure independent semantic encoding. Formulated as:
\begin{equation}
    M_{i,j}^{(intra)} = 1 \iff \exists u \in \mathcal{U} \text{ s.t. } \{i, j\} \subseteq \mathcal{T}(u) \land j \le i
\end{equation}
This design aligns perfectly with the LM's pre-training paradigm, allowing the model to ``inherit'' its native linguistic capability to encode local semantics \textit{as if processing independent samples in a batch.}

\textbf{Level 2: Inter-Element Structural Flow.}

To enable structural reasoning akin to GNNs, we designate the \textit{closing tags} (\texttt{</n>}, \texttt{</e>}) as \textbf{Semantic Hubs}. Since they naturally aggregate complete internal semantics via the causal mask (Level 1), they serve as ideal communication ports for transmitting encapsulated information to neighbors.

We model the topology as a directed attention flow chain: \textit{Source Node $\to$ Edge $\to$ Target Node}. For a triplet $v_{src} \xrightarrow{e} v_{tgt}$, we establish the connectivity logic as:
\begin{equation}
\begin{split}
    M_{i,j}^{(inter)} = 1 \iff & \exists (v_{src} \xrightarrow{e} v_{tgt}) \text{ s.t. } \\
    & (i=\text{hub}(e) \land j=\text{hub}(v_{src})) \lor \\
    & (i=\text{hub}(v_{tgt}) \land j=\text{hub}(e))
\end{split}
\end{equation}
This creates a differentiable message-passing path where information flows sequentially through the topology. For undirected edges, this logic is applied bidirectionally.

\textbf{Level 3: Global Connectivity.}
The global tags orchestrate the holistic graph representation. The closing tag \texttt{</g>} acts as a virtual super node~\cite{pham2017graph} that aggregates information from all element Hubs. Conversely, the start tag \texttt{<g>} serves as a universal anchor visible to all other tokens.
\begin{equation}
\begin{split}
M^{(global)}_{i, j} = 1 & \iff \bigl(i = \text{hub}(\mathcal{G}) \land j \in \\ & \{ \text{hub}(u) \mid u \in \mathcal{U} \}\bigr) \lor \; j = \text{start}(\mathcal{G})
\end{split}
\end{equation}
where $\text{start}(\mathcal{G})$ denotes the index of \texttt{<g>}.

\textbf{Level 4: Query-Graph Interaction.}
Finally, the query tokens $Q$ must perform autoregressive generation while retrieving information from the graph. Therefore, a query token $i \in Q$ must attend to both its preceding query tokens (standard causal masking) and specific graph elements. We explore two strategies:
\begin{itemize}
    \setlength\itemsep{0em}
    \item \textbf{Sparse Attention:} $Q$ attends to previous query tokens and Semantic Hubs. This is computationally efficient:
    \begin{equation}
    \begin{split}
        M_{i,j}^{(query)} = 1 & \iff  (i, j \in Q \land j \le i) \lor \\
        & (i \in Q \land j \in \{ \text{hub}(u) \mid u \in \mathcal{U} \})
    \end{split}
    \end{equation}

    \item \textbf{Full Attention:} $Q$ attends to previous query tokens and all graph tokens. This maximizes context capacity:
    \begin{equation}
    M_{i,j}^{(query)} = 1 \iff i \in Q \land j \le i
    \end{equation}
\end{itemize}
We provide an empirical analysis and discussion of the trade-offs between these two strategies in Section \ref{sec:analysis_strategy}.

\textbf{Unified Mask Composition.}
The final topology-aware mask $M$ is constructed by the logical disjunction of the visibility constraints defined at all four levels:
\begin{equation}
    M_{i,j} = M_{i,j}^{(intra)} \lor M_{i,j}^{(inter)} \lor M_{i,j}^{(global)} \lor M_{i,j}^{(query)}
\end{equation}
This ensures that the attention mechanism simultaneously respects local semantic integrity, topological connectivity, global aggregation, and query content.

% =====================================

\subsection{Structural Position Calibration}
\label{sec:position}

While the Flattening strategy (Section \ref{sec:input_construction}) allows graphs to be processed as sequences, assigning standard monotonically increasing positional IDs (i.e., $0, 1, 2, 3, \dots$), which is the ubiquitous indexing scheme in contemporary LMs, introduces a spurious sequential bias: the relative distance between the query tokens $Q$ and different graph elements varies based on their arbitrary input order. To enforce \textbf{structural equivalence}, we leverage Rotary Positional Embeddings (RoPE)~\cite{su2024roformer}, widely used in modern LMs. Crucially, RoPE encodes positions such that attention scores depend solely on the relative distance between tokens.

Based on this property, we propose a \textbf{Recalibrated Indexing Strategy} to ensure topologically parallel elements are equidistant to the query. As illustrated in Figure \ref{fig:main}, instead of continuous incrementation, we synchronize the positional IDs of all Semantic Hubs. Specifically, while IDs within each element increment naturally to preserve internal token order, the closing tags (Hubs) of \textit{all} elements are forced to share a unified positional ID:
\begin{equation}
p_{hub} = p_{start} + \max_{u \in \mathcal{U}}(|u|) 
\end{equation}
where $p_{start}$ denotes the positional ID of the graph start tag \texttt{<g>}. Subsequently, the graph end tag \texttt{</g>} is assigned $p_{hub} + 1$, and query tokens resume standard incremental indexing. By anchoring all Hubs to $p_{hub}$, every graph element presents a uniform structural distance to $Q$. 

This design \textit{mathematically guarantees} that the attention mechanism is \textit{invariant to the serialization order of graph elements}, ensuring that reasoning is driven purely by semantics and topology regardless of input permutation.

% =====================================
\subsection{Efficient Training Implementations}
\label{sec:training}

\begin{figure}[t]
  % \vskip 0.2in
  \begin{center}
  \centerline{\includegraphics[width=0.6\columnwidth]{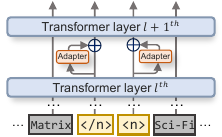}}
    \caption{
    Illustration of the \themodel-Zero Mechanism. Residual adapters operate exclusively on special tokens, while semantic text tokens pass through unchanged. This selective processing guarantees zero interference with the backbone LM.
    }
    \label{fig:adapter}
  \end{center}
\end{figure}

To adapt the LM to the text-graph domain, we propose two distinct training strategies. In both configurations, the embeddings for the newly introduced special tokens (i.e., \texttt{<g>}, \texttt{<n>}, \texttt{<e>}, etc.) are added to the vocabulary and optimized.

\textbf{NAG-LoRA.}
For the first strategy, we employ Low-Rank Adaptation (LoRA)~\cite{hu2021lora}. We inject trainable rank-decomposition matrices into the query, key, and value projections ($W_Q, W_K, W_V$) of the LM's attention mechanism. This allows the model to adjust its internal attention dynamics to fully exploit the topology-aware mask.

\textbf{NAG-Zero.}
For the second strategy, aimed at ensuring \textbf{zero interference} with the original LM's linguistic abilities, we design a lightweight mechanism that operates between frozen Transformer layers (see Figure \ref{fig:adapter}). Crucially, this adapter is activated \textit{exclusively} for special structural tokens, while standard text tokens bypass the module entirely. Consequently, for plain text inputs devoid of graph structures, the model functions identically to the original backbone, thereby guaranteeing absolute preservation of general capabilities—hence the designation ``Zero.''

Formally, let $h^{(l)}$ be the hidden state of a special token output by layer $l$. The input to the subsequent layer $l+1$ is computed via a gated residual transformation:
\begin{equation}
\begin{aligned}
    \tilde{h}^{(l)} &= h^{(l)} + \text{Adapter}(h^{(l)})\\
    \text{Adapter}(h^{(l)}) &= \sigma(W_{g}^{U} W_{g}^{D}h^{(l)}) \odot (W_{v}^{U} W_{v}^{D}h^{(l)})
\end{aligned}
\end{equation}
where both the gating projection (subscript $g$) and the value projection (subscript $v$) are parameterized as low-rank bottlenecks ($W^{D} \in \mathbb{R}^{r \times d}, W^{U} \in \mathbb{R}^{d \times r}$ with $r < d$) to minimize parameter overhead, and $\sigma(\cdot)$ is the sigmoid function. 

%=================================

\subsection{Theoretical Intuition: Tuning the Message Function}
\label{sec:theory}

We justify the design of \themodel by formally bridging it with the Message Passing GNN paradigm. Typically, node representations are updated by aggregating messages from the neighborhood $\mathcal{N}(i) \cup \{i\}$:
\begin{equation}
    h_i^{new} = \text{Agg}\left( \left\{ \phi(h_j) \mid j \in \mathcal{N}(i) \cup \{i\} \right\} \right)
    \label{eq:gnn_agg}
\end{equation}
where $\phi(\cdot)$ is a learnable message function and $\text{Agg}$ is an aggregation operator (e.g., weighted average).

Our Topology-Aware Mask $M$ strictly enforces this neighborhood within the self-attention mechanism. For a single attention head in \themodel-Zero, the updated representation for hub $i$ is a weighted aggregation:
\begin{equation}
    h_i^{new} = \sum_{j: M_{i,j}=1} \alpha_{ij} \left(W_V \cdot \left(h_j+\text{Adapter}(h_j)\right)\right)
\end{equation}
where $\alpha_{ij}$ is the attention score and $W_V$ is the LM's intrinsic value projection matrix. By aligning this with Eq.~\ref{eq:gnn_agg}, we identify the effective message function $\phi$ as the composition:
\begin{equation}
    \phi(h) = W_V \cdot \left(h+\text{Adapter}(h)\right)
\end{equation}
Consequently, optimizing the \themodel-Zero parameters is mathematically equivalent to tuning the message function $\phi$ directly within the Transformer's residual stream.
\begin{table*}[t]
\centering
\caption{Experimental Results on Graph, Node, and Edge Reasoning Tasks. Accuracy (\%) is reported as the primary metric for all tasks, with Absolute Error ($\downarrow$) or F1-score ($\uparrow$) provided in parentheses for regression and connectivity tasks. \textbf{Bold} indicates the best performance.}
\label{tab:topo_graph_results}
\small
\setlength{\tabcolsep}{0pt}
\renewcommand{\arraystretch}{1.2}
\begin{tabular*}{\textwidth}{@{\extracolsep{\fill}}lccccccccc}
\toprule
& \multicolumn{4}{c}{\textbf{Graph-level Tasks}} & \multicolumn{2}{c}{\textbf{Node-level Tasks}} & \multicolumn{3}{c}{\textbf{Edge-level Tasks}} \\
\cmidrule(lr){2-5} \cmidrule(lr){6-7} \cmidrule(lr){8-10}
\textbf{Method} & \begin{tabular}[c]{@{}c@{}}Node\\ Count\\ \textit{\scriptsize Acc$\uparrow$ (Abs Err$\downarrow$)}\end{tabular} & \begin{tabular}[c]{@{}c@{}}Edge\\ Count\\ \textit{\scriptsize Acc$\uparrow$ (Abs Err$\downarrow$)}\end{tabular} & \begin{tabular}[c]{@{}c@{}}Cycle\\ Check\\ \textit{\scriptsize Acc$\uparrow$ }\end{tabular} & \begin{tabular}[c]{@{}c@{}}Triangle\\ Count\\ \textit{\scriptsize Acc$\uparrow$ (Abs Err$\downarrow$)}\end{tabular} & \begin{tabular}[c]{@{}c@{}}Node\\ Degree\\ \textit{\scriptsize Acc$\uparrow$ (Abs Err$\downarrow$)}\end{tabular} & \begin{tabular}[c]{@{}c@{}}Connected\\ Nodes\\ \textit{\scriptsize Acc$\uparrow$ (F1$\uparrow$)}\end{tabular} & \begin{tabular}[c]{@{}c@{}}Reach-\\ ability\\ \textit{\scriptsize Acc$\uparrow$ }\end{tabular} & \begin{tabular}[c]{@{}c@{}}Edge\\ Existence\\ \textit{\scriptsize Acc$\uparrow$ }\end{tabular} & \begin{tabular}[c]{@{}c@{}}Shortest\\ Path\\ \textit{\scriptsize Acc$\uparrow$ (Abs Err$\downarrow$)}\end{tabular} \\
\midrule
Qwen3-Direct & 45.65 {\scriptsize (3.81)} & 31.15 {\scriptsize (35.21)} & 76.20  & 11.15 {\scriptsize (149.38)} & 57.00 {\scriptsize (1.35)} & 15.60 {\scriptsize (0.65)} & 98.70  & 74.30  & 37.25 {\scriptsize (0.81)} \\ %this is 8B
$\text{GraphToken}_\text{G}$ & 82.02 {\scriptsize (0.19)} & 40.60 {\scriptsize (2.25)} & 97.95  & 50.88 {\scriptsize (12.38)} & 50.67 {\scriptsize (1.14)} & 35.62 {\scriptsize (0.69)} & 97.59  & 78.65  & 62.21 {\scriptsize (0.66)} \\
$\text{GraphToken}_\text{N}$ & 94.38 {\scriptsize (0.06)} & 45.97 {\scriptsize (2.16)} & 96.92  & 56.54 {\scriptsize (12.99)} & 52.01 {\scriptsize (1.09)} & 33.90 {\scriptsize (0.69)} & 97.59  & 78.64  & 61.89 {\scriptsize (0.67)} \\
$\text{GraphToken}_\text{E}$ & 93.26 {\scriptsize (0.13)} & 59.73 {\scriptsize (1.50)} & 97.95  & 65.02 {\scriptsize (4.49)} & 60.74 {\scriptsize (0.85)} & 38.70 {\scriptsize (0.70)} & 97.80  & 77.58  & 61.89 {\scriptsize (0.56)} \\
Qwen3-LoRA & \textbf{100.00} {\scriptsize (0.00)} & 86.35 {\scriptsize (0.23)} & 99.80  & 68.00 {\scriptsize (2.04)} & 93.25 {\scriptsize (0.07)} & \textbf{96.85} {\scriptsize (1.00)} & 99.45  & 99.55  & 91.25 {\scriptsize (0.10)} \\
\midrule
\themodel-Zero (Ours) & 99.85 {\scriptsize (0.00)} & 93.00 {\scriptsize (0.09)} & 99.35  & 64.95 {\scriptsize (1.62)} & 66.21 {\scriptsize (0.64)} & 52.40 {\scriptsize (0.80)} & 97.80  & 80.70  & 63.40 {\scriptsize (0.77)} \\
\themodel-LoRA (Ours) & \textbf{100.00} {\scriptsize (0.00)} & \textbf{94.95} {\scriptsize (0.06)} & \textbf{99.90}  & \textbf{74.35} {\scriptsize (0.89)} & \textbf{99.75} {\scriptsize (0.00)} & 84.90 {\scriptsize (0.98)} & \textbf{99.90}  & \textbf{99.70}  & \textbf{95.00} {\scriptsize (0.06)} \\
\bottomrule
\end{tabular*}
\end{table*}

\section{Experiments}
\label{sec:experiments}

We evaluate the \themodel framework via a two-stage strategy that examines both structural perception and semantic reasoning:

\begin{itemize}
\setlength{\itemsep}{0pt}
    \item \textbf{Topological Awareness}: This stage focuses on the model's intrinsic ability to perceive graph connectivity through a suite of topology-centric tasks.
    \item \textbf{Semantic Graph Reasoning}: This stage emphasizes the integration of topological understanding with linguistic knowledge to solve reasoning problems in real-world scenarios.
\end{itemize}

% ==============================================

\subsection{Experimental Setup and Baselines}
\label{sec:setup}

We utilize Qwen3~\cite{yang2025qwen3} as our backbone language model. We compare \themodel\ against representative baselines from two distinct paradigms.

The first category encompasses linearization-based baselines, where graphs are serialized into textual triplets as established by \citet{fatemi2023talklikegraphencoding} (see Appendix \ref{sec:appendix_graph_linear} for details). Utilizing this format, we evaluate Qwen3-Direct in a zero-shot setting and Qwen3-LoRA via standard LoRA fine-tuning.

The second category represents dual-path architectures (GNN+LM). We implement an adapted version of GraphToken~\cite{perozzi2024let} that integrates GNN-encoded soft prompts with explicit graph textual descriptions~\cite{he2024g}. We evaluate three structural granularities: $\text{GraphToken}_\text{G}$, which integrates a global pooled representation (architecturally equivalent to G-Retriever~\cite{he2024g}), and its fine-grained counterparts $\text{GraphToken}_\text{N}$ and $\text{GraphToken}_\text{E}$, which inject embeddings at the node and edge levels, respectively. Detailed implementation specifications for all three variants are provided in Appendix \ref{sec:appendix_dual_path}. To assess the capability of capturing graph topology within accessible budgets, we validate the proposed framework using small-scale LMs; all experiments can be implemented on a single NVIDIA RTX 4090 GPU.

% ===============================================

\subsection{Topological Awareness}
\label{sec:topo_aware}

\subsubsection{Datasets and Metrics}
Following the protocol established by \citet{fatemi2023talklikegraphencoding}, we construct a comprehensive synthetic benchmark to evaluate intrinsic graph understanding. To ensure topological diversity, graphs are generated using diverse algorithms, including Scale-Free Networks~\cite{barabasi1999emergence}, Stochastic Block Models~\cite{holland1983stochastic}, and Erd\H{o}s-R\'enyi graphs~\cite{erdds1959random}, among others.
We randomize node identifiers using diverse schemes (e.g., random names, integer indices, or alphabets) and vary the textual descriptions of edge relations (e.g., utilizing terms like ``friendship'', ``connection''). All graphs utilize undirected edges to test bidirectional reasoning. Further generation details are provided in Appendix \ref{sec:appendix_topo_dataset}.

The benchmark encompasses 9 tasks across three levels of granularity: (1) \textit{Graph-level:} Node Count, Edge Count, Cycle Check, Triangle Count. (2) \textit{Node-level:} Node Degree, Connected Nodes (neighbor retrieval). (3) \textit{Edge-level:} Reachability, Edge Existence, Shortest Path.

\textbf{Evaluation Metrics.}
We employ accuracy as the primary metric. For regression tasks (e.g., counting, shortest path length), we additionally report the absolute error (Abs Err). For the \textit{Connected Nodes} task, which requires listing the textual content of all adjacent nodes for a given query, we report the F1-score. This metric comprehensively accounts for both hallucinations and omissions, providing a more robust measure than simple accuracy for set retrieval.
Unless otherwise stated, all models are based on the compact 600M parameter version of Qwen3. The Qwen3-Direct baseline utilizes the larger 8B version solely for reference purposes.

\subsubsection{Results Analysis}
The most critical insight from Table~\ref{tab:topo_graph_results} stems from the information input modality. Unlike linearization baselines that explicitly describe connectivity via text triplets, or GraphToken methods that inject pre-computed structural features, \themodel\ perceives topology \textit{solely} through the topology-aware attention mechanism. It receives no explicit textual descriptions of edge connections. The superior performance across most tasks confirms that this injection is highly effective: the Semantic Hubs successfully capture and propagate topological signals purely via attention masking.

\textbf{Efficacy of Non-Invasive Structure Injection.}
\themodel-Zero, which only trains lightweight adapters for special tokens, exhibits significant superiority over \textit{all} parameterized GraphToken variants. While GraphToken shows sensitivity to embedding granularity (e.g., $\text{GraphToken}_\text{G}$ underperforms on \textit{Edge Count} while $\text{GraphToken}_\text{N}$ struggles with \textit{Connected Nodes}), \themodel-Zero achieves robust and consistent performance across the board. This suggests that constraining the attention flow is a more effective inductive bias for structure than soft prompt injection.

\textbf{Superiority in Structural Reasoning.}
\themodel-LoRA demonstrates dominant performance, achieving the best results in 8 out of 9 tasks. Notably, on computationally intensive geometric tasks like \textit{Triangle Count} and \textit{Shortest Path}, \themodel-LoRA significantly outperforms the text-only Qwen3-LoRA (Triangle Count Acc: 74.35\% vs. 68.00\%; Shortest Path Abs Err: 0.06 vs. 0.10). This indicates that standard LMs face limitations when handling complex spatial reasoning, a gap that our topology-aware mask effectively bridges. 
However, we acknowledge a performance gap in the \textit{Connected Nodes} task, where Qwen3-LoRA outperforms \themodel-LoRA (F1: 1.00 vs. 0.98). Error analysis reveals that our model tends to omit items specifically when a node has a high degree. This suggests that the Semantic Hubs face an information bottleneck when aggregating extensive neighborhood contexts, whereas linearization-based methods benefit from explicitly presenting all neighbor names in the text sequence for the LM to retrieve.

% =============================================

\begin{table}[t]
\centering
\caption{Experimental Results on Semantic Graph Reasoning Datasets. Accuracy (\%) is reported for ExplaGraphs and SceneGraphs, while Hit@1 (\%) is reported for WebQSP. \textbf{Bold} indicates the best performance.}
\label{tab:semantic_results}
\small
\setlength{\tabcolsep}{0pt}
\renewcommand{\arraystretch}{1.2}
\begin{tabular*}{\columnwidth}{@{\extracolsep{\fill}}lccc}
\toprule
\textbf{Method} & \textbf{ExplaGraphs} & \textbf{SceneGraphs} & \textbf{WebQSP} \\
\midrule
Qwen3-Direct & 58.84 & 50.00 & 32.46 \\
Qwen3-LoRA & 62.09 & 83.71 & 44.37 \\
$\text{GraphToken}_\text{G}$  & 65.52 & 68.37 & 40.56 \\
$\text{GraphToken}_\text{N}$  & 70.94 & 73.65 & 41.41 \\
$\text{GraphToken}_\text{E}$  & 80.51 & 72.66 & 42.39 \\
\midrule
NAG-Zero & 78.16 & 71.51 & 41.40 \\
NAG-LoRA & \textbf{82.49} & \textbf{83.82} & \textbf{55.25}  \\
\bottomrule
\end{tabular*}
\end{table}

\subsection{Semantic Graph Reasoning}
\label{sec:semantic_reasoning}

\subsubsection{Datasets and Metrics}
Following the experimental settings in~\cite{he2024g}, we evaluate our framework on three representative benchmarks that require integrating structural topology with diverse semantics.
ExplaGraphs is a commonsense reasoning dataset for debate stance prediction, requiring the model to assess whether a structured argument supports or counters a belief.
SceneGraphs evaluates visual-spatial reasoning, where the model must answer open-ended questions based on scene graph descriptions containing objects, attributes, and spatial relations.
WebQSP is a large-scale Knowledge Base QA dataset based on Freebase, challenging the model to perform multi-hop reasoning over entity facts. For SceneGraphs and WebQSP, we utilize query-retrieved subgraphs as the input graphs, as detailed in Appendix \ref{sec:appendix_semantic_dataset}.

\textbf{Evaluation Metrics.}
We report Accuracy for ExplaGraphs and SceneGraphs. For WebQSP, given the potential for multiple valid entities, we utilize Hit@1 to measure the precision of the top predicted answer. All experiments are conducted using the 600M parameter version of Qwen3 as the backbone.

\subsubsection{Results Analysis}
Table~\ref{tab:semantic_results} summarizes the performance across real-world semantic graph benchmarks. In contrast to the synthetic topological tasks, these datasets demand a rigorous synthesis of both structural connectivity and rich textual semantics.

\textbf{Superiority of \themodel-LoRA.}
\themodel-LoRA consistently achieves the best performance across all three datasets. On ExplaGraphs, it reaches an accuracy of 82.49\%, surpassing the GraphToken baseline (80.51\%) and showing a distinct advantage over the linearization-based Qwen3-LoRA (62.09\%). Similarly, on WebQSP, it leads with 55.25\% compared to 44.37\% for the standard LoRA baseline. This confirms that even in semantically rich scenarios, explicit topological modeling provides a tangible gain over sequential text processing.

\textbf{Capacity Gap in Semantic Tasks.}
A key observation is the pronounced performance gap between \themodel-LoRA and \themodel-Zero, which is wider here than in the pure topological tasks. This trend parallels the results on the \textit{Connected Nodes} task, where heavy reliance on textual content enumeration widened the gap. It suggests that while lightweight adapters in \themodel-Zero effectively capture intrinsic topology, tasks necessitating deep semantic integration face a performance ceiling due to the restricted scope of inter-layer modulation. Capturing complex semantic nuances requires fine-tuning a broader range of model parameters, as achieved by LoRA's direct adjustment of all internal attention weights.

\textbf{Comparison with Dual-Path Baselines.}
When comparing \themodel-Zero against the GraphToken variants, which also aim to inject structure without full fine-tuning, the results are largely comparable. For instance, on ExplaGraphs, \themodel-Zero (78.16\%) performs on par with the GraphToken methods (ranging from 65.5\% to 80.5\%). Considering that GraphToken baselines explicitly encode graph features into the input sequence, \themodel-Zero demonstrates that a purely attention-based structural injection remains a competitive and parameter-efficient alternative.

% =======================================
\begin{figure}[t]
  \vskip 0.2in
  \begin{center}
  \centerline{\includegraphics[width=1\columnwidth]{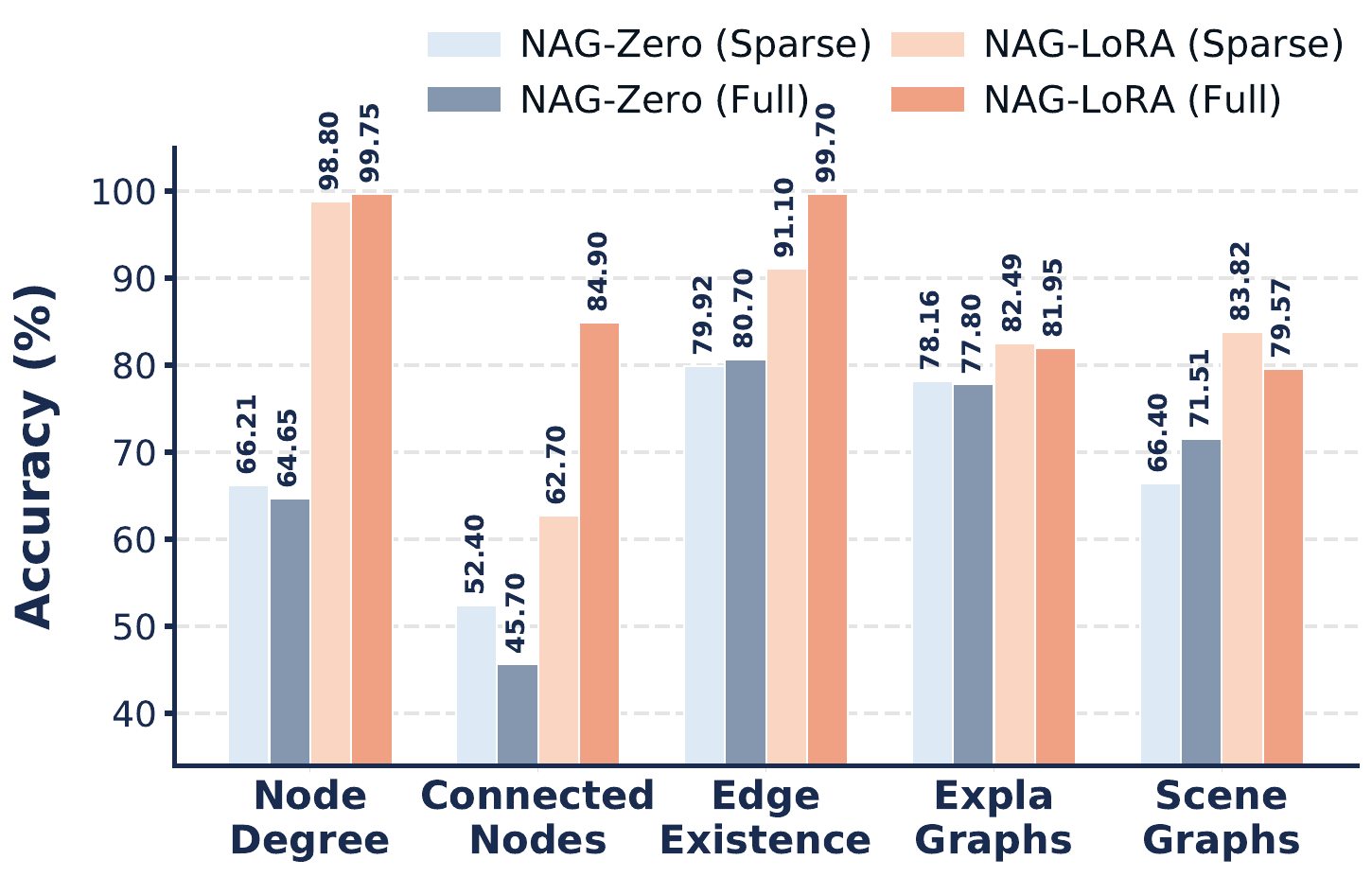}}
    \caption{
    Analysis of query-graph interaction strategies. We illustrate the performance trade-offs between \textit{Sparse} and \textit{Full} attention mechanisms under both NAG-Zero (blue) and NAG-LoRA (orange) settings.}
    \label{fig:interaction}
  \end{center}
\end{figure}

\subsection{Analysis of Interaction Strategies}
\label{sec:analysis_strategy}

Figure~\ref{fig:interaction} illustrates the performance dynamics between \textit{Sparse} and \textit{Full} attention mechanisms. Crucially, we observe no consistent dominance of one strategy over the other; the preference varies not only by task but also shifts distinctively between Zero and LoRA configurations (e.g., in SceneGraphs, the optimal strategy reverses across training regimes). This fluctuation highlights an inherent trade-off between the noise reduction offered by sparsity and the comprehensive global context provided by full connectivity. Such variability indicates that static patterns are insufficient. Instead, effective structural injection requires a dynamic mechanism that modulates attention flows based on the specific interaction between the input query and the underlying graph topology.

\begin{table}[t]
\centering
\caption{Ablation Study on Positional Indexing Strategies. Experiments are conducted using the NAG-LoRA (Sparse) configuration. Accuracy (\%) is reported for all tasks. \textbf{Bold} indicates the best performance.}
\label{tab:ablation_position}
\small
\setlength{\tabcolsep}{3pt} 
\renewcommand{\arraystretch}{1.2}
\begin{tabular*}{\columnwidth}{@{\extracolsep{\fill}}lccccc}
\toprule
\textbf{Strategy} & \begin{tabular}[c]{@{}c@{}}\textbf{Node}\\\textbf{Degree}\end{tabular} & \begin{tabular}[c]{@{}c@{}}\textbf{Conn.}\\\textbf{Nodes}\end{tabular} & \begin{tabular}[c]{@{}c@{}}\textbf{Edge}\\\textbf{Exist.}\end{tabular} & \begin{tabular}[c]{@{}c@{}}\textbf{Expla}\\\textbf{Graphs}\end{tabular} & \begin{tabular}[c]{@{}c@{}}\textbf{Scene}\\\textbf{Graphs}\end{tabular} \\
\midrule
Sequential Pos. & 98.45 & 59.75 & 90.15 & \textbf{83.39} & 82.99 \\
\textbf{Recalibrated Pos.} & \textbf{98.80} & \textbf{62.70} & \textbf{91.10} & 82.49 & \textbf{83.82} \\
\bottomrule
\end{tabular*}
\end{table}

\subsection{Impact of Structural Position Calibration}
\label{sec:ablation_pos}

Finally, we validate the hypothesis that standard sequential indexing introduces arbitrary order bias that hinders graph perception. We conduct an ablation study by replacing our Recalibrated Indexing Strategy (which assigns equivalent IDs to parallel Semantic Hubs) with standard sequential indexing (where IDs increment linearly following the token sequence). As shown in Table~\ref{tab:ablation_position}, adopting Recalibrated Indexing consistently boosts performance across the majority of semantic and topological tasks.

The most significant gain is observed in the \textit{Connected Nodes} task (+2.95\%). This substantial improvement confirms that enforcing structural equivalence through positional IDs effectively mitigates the bias introduced by arbitrary linearization. By placing topologically parallel nodes on the same positional plane, the model avoids inferring spurious hierarchical relationships from the serialization order, leading to more robust structural comprehension.

\section{Discussion and Future Work}
\label{sec:discussion}

Our investigation highlights three key avenues for future research.
\textit{First}, the comparable performance of Sparse and Full interaction modes suggests that static attention patterns may be suboptimal for diverse query complexities. Future work should explore dynamic sparsity mechanisms, enabling the model to adaptively modulate attention focus based on the interplay between the query intent and the graph topology.
\textit{Second}, the performance gap between \themodel-Zero and LoRA on semantic tasks reveals the limited modulation space of standard adapters. This necessitates more expressive parameter-efficient designs to achieve deeper semantic alignment without compromising the backbone LM's inherent linguistic capabilities.
\textit{Finally}, our resource-constrained focus on compact LMs leaves the scalability of topological awareness to larger foundation models and complex, dynamic graph scenarios as a critical open question for verifying the robustness and scalability of our approach.

\section{Conclusion}
\label{sec:conclusion}

In this work, we introduced \themodel, a framework that enables Language Models to natively perceive and reason over graph structures. By integrating a topology-aware attention mechanism with structural position calibration, we successfully eliminate the reliance on external graph encoders. Extensive experiments confirm that \themodel not only masters fundamental topological tasks but also effectively facilitates semantic reasoning in real-world scenarios. Our findings suggest a paradigm shift: structural understanding can be intrinsically unified within the Transformer architecture, offering a streamlined, encoder-free alternative for graph-text interaction.
%================================================================

% In the unusual situation where you want a paper to appear in the
% references without citing it in the main text, use \nocite
% \nocite{langley00}

\bibliography{mybib}
\bibliographystyle{icml2026}

%%%%%%%%%%%%%%%%%%%%%%%%%%%%%%%%%%%%%%%%%%%%%%%%%%%%%%%%%%%%%%%%%%%%%%%%%%%%%%%
%%%%%%%%%%%%%%%%%%%%%%%%%%%%%%%%%%%%%%%%%%%%%%%%%%%%%%%%%%%%%%%%%%%%%%%%%%%%%%%
% APPENDIX 
%%%%%%%%%%%%%%%%%%%%%%%%%%%%%%%%%%%%%%%%%%%%%%%%%%%%%%%%%%%%%%%%%%%%%%%%%%%%%%%
%%%%%%%%%%%%%%%%%%%%%%%%%%%%%%%%%%%%%%%%%%%%%%%%%%%%%%%%%%%%%%%%%%%%%%%%%%%%%%%
\newpage
\appendix
\onecolumn
\section{Dataset Settings}
\label{sec:appendix_dataset}
\subsection{Topological Reasoning Dataset}
\label{sec:appendix_topo_dataset}
To construct the dataset for topological tasks, we follow the data pipeline proposed by \cite{fatemi2023talklikegraphencoding} to generate data with diverse graph structures, text encodings, and reasoning tasks. We utilize the NetworkX library~\cite{hagberg2007exploring} to generate graphs via standard algorithms, encompassing seven classic topologies: Erd\H{o}s-R\'enyi (ER)~\cite{erdds1959random}, Barabási-Albert (BA) scale-free networks~\cite{barabasi1999emergence}, Stochastic Block Models~\cite{holland1983stochastic}, Watts-Strogatz (WS) small-world networks~\cite{watts1998collective}, Complete graphs, Star graphs, and Path graphs. To balance structural diversity with the limited context window of language models, the graph sizes are constrained to 5–20 nodes and 0-200 edges, encompassing varying sparsity levels. Following the original pipeline, we employ seven distinct text encoding methods (i.e., serialization schemes that map nodes and edges to diverse textual identifiers and relationship descriptions): \textit{adjacency}, \textit{expert}, \textit{friendship}, \textit{game of thrones}, \textit{politician}, \textit{social network}, and \textit{south park}. To comprehensively evaluate the model's ability on topological structural understanding, we test nine tasks as follows:

\begin{itemize}
\item \textbf{Node Count:} Count the total number of nodes in a graph.
\item \textbf{Edge Count:} Count the total number of edges in a graph.
\item \textbf{Cycle Check:} Determine whether a graph contains at least one cycle.
\item \textbf{Triangle Count:} Count the total number of triangles present in the graph.
\item \textbf{Node Degree:} Calculate the degree of a specified node.
\item \textbf{Connected Nodes:} Identify all nodes that are directly connected to a given node.
\item \textbf{Reachability:} Determine whether a valid path exists between two specified nodes.
\item \textbf{Edge Existence:} Determine whether a specific edge exists between two nodes.
\item \textbf{Shortest Path:} Calculate the length of the shortest path between two specified nodes.
\end{itemize}

To ensure data richness and robustness, we perform an equal proportion mixture on graph structure and text encoding settings. By sampling 20,000 within each task, we obtained a comprehensive dataset totaling $20,000 \times 9 = 180,000$ samples. Table ~\ref{tab:datasets_summary_topo} presents the statistical details of the final dataset. In the training process, the dataset is then divided into train/validation/test splits at 8:1:1 ratio.

\begin{table}[h]
\centering
\caption{Statistical Details of Topological Reasoning Dataset}
\label{tab:datasets_summary_topo}
\small
\renewcommand{\arraystretch}{1.2}
\begin{tabular}{lccc} 
\toprule
\textbf{Dataset} & \textbf{GraphQA} \\
\midrule
\#Graphs & 180,000 \\
Avg. \#Nodes & 12.11 \\
Avg. \#Edges & 31.42 \\
Task & Topological Reasoning \\
Evaluation Metrics & Accuracy \\
\bottomrule
\end{tabular}
\end{table}

\subsection{Semantic Graph Reasoning Dataset}
\label{sec:appendix_semantic_dataset}

To construct the semantic graph reasoning dataset, we leverage the benchmark suite from G-Retriever~\cite{he2024g}, which standardizes and processes three well-established datasets: ExplaGraph~\cite{explagraphs}, SceneGraph~\cite{scenegraph}, and WebQSP~\cite{webqsp,webqspfollowing}, to evaluate model performance in cross-domain graph reasoning tasks. Specifically, these datasets impose distinct reasoning demands: ExplaGraphs assesses generative commonsense reasoning within argumentation structures; SceneGraphs highlights the complexity of rich node attributes (e.g., color, shape), where questions transcend the simple verification of node existence or names; and WebQSP targets knowledge-based QA in real-world scenarios, explicitly requiring multi-hop reasoning to derive correct answers. We maintain the subgraph extraction protocols and splitting ratios, but exclude the non-participating coordinate information from the SceneGraph data to optimize the model's input context length. The statistical profiles of the processed datasets are summarized in Table~\ref{tab:datasets_summary_semantic}.

\begin{table}[h]
\centering
\caption{Statistical Details of Semantic Graph Reasoning Dataset after Subgraph Extraction}
\label{tab:datasets_summary_semantic}
\small
\setlength{\tabcolsep}{0pt}
\renewcommand{\arraystretch}{1.2}
\begin{tabular*}{\columnwidth}{@{\extracolsep{\fill}}lccc}
\toprule
\textbf{Dataset} & \textbf{ExplaGraphs} & \textbf{SceneGraphs} & \textbf{WebQSP} \\
\midrule
\#Graphs & 2,766 & 99,230 & 4,517 \\
Avg. \#Nodes & 5.17 & 8.77 & 11.77 \\
Avg. \#Edges & 4.25 & 13.41 & 11.86 \\
Task & Common sense reasoning & Scene graph question answering & Knowledge based question answering \\
Evaluation Metrics & Accuracy & Accuracy & Hit@1 \\
\bottomrule
\end{tabular*}
\end{table}

\section{Baseline Implementation Details}
\label{sec:appendix_baseline}

\subsection{Graph Linearization}
\label{sec:appendix_graph_linear}

To evaluate the capabilities of standard LMs on graph reasoning tasks without modifications, we employ a textual linearization strategy. We adhere to distinct linearization schemes for topological and semantic tasks to align with the specific data characteristics and prior baselines.

\paragraph{Linearization Scheme I: Topological Reasoning.}
For the synthetic topological benchmark, we adopt a \textit{Text-Tuple} serialization method. This approach explicitly serializes the graph $\mathcal{G}$ into a narrative description comprising a node set listing and a sequence of edge triplets.

\begin{itemize}
    \item \textbf{Header Definition:} A fixed prefatory sentence defines the semantics of the tuple $(s, p, o)$, establishing that $s$ and $o$ are nodes connected by an edge of type $p$.
    \item \textbf{Node \& Edge Lists:} The prompt first enumerates all node identifiers in $\mathcal{V}$, followed by the edge set $\mathcal{E}$ presented as parenthesized triplets. The specific naming convention is determined by the textual encoding method of the data item (e.g., \texttt{(A, edge, B), (John, coauthor, Joseph)}).
\end{itemize}

\paragraph{Prompt Template I}
The prompt construction follows a strict template to facilitate zero-shot evaluation. The format is defined as follows:

\begin{center}
\fbox{
    \begin{minipage}{0.95\linewidth}
        \small
        \textbf{Graph Description:}\\
        In an undirected graph, ($s, p, o$) means that node $s$ and node $o$ are connected with an undirected edge of type $p$.\\ G describes a graph among nodes: $\{Node List\}$\\
        The edges in G are: $\{Edge List\}$\\
        \textbf{Question:}\\
        $\{Question\}$
        
        \textbf{Constraint:}\\
        Please provide a direct answer to the question.
        
        \textbf{Answer:}
    \end{minipage}
}
\end{center}

\paragraph{Linearization Scheme II: Semantic Reasoning.}
For real-world semantic datasets (\textit{ExplaGraphs}, \textit{SceneGraphs}, \textit{WebQSP}), we utilize a \textit{CSV-style Node-Edge List} linearization as in \citet{he2024g}. This format is designed to handle rich textual attributes associated with nodes (e.g., argument text or object descriptions) by separating structural definitions into indexed CSV blocks.
\begin{itemize}
    \item \textbf{Node Block:} A list formatted as \texttt{node\_id, "node\_attr"}, mapping numerical indices to textual content.
    \item \textbf{Edge Block:} A list formatted as \texttt{src, "edge\_attr", dst}, defining directed connections between indices.
\end{itemize}

\paragraph{Prompt Template II}
The template for semantic tasks incorporates these CSV blocks within a definition wrapper:

\begin{center}
\fbox{
    \begin{minipage}{0.95\linewidth}
        \small
        \textbf{Graph Description:}\\
        In a directed graph G:\\
        node\_id, node\_attr means the node index and its attribute in the graph.\\
        example: 1, ``attribute text"\\
        src, edge\_attr, dst means that node src and node dst are connected with a directed edge of type edge\_attr.\\
        example: 5, ``relation", 10
        
        The nodes, and its attributes in G are:\\
        \texttt{node\_id,node\_attr}\\
        \textit{0, ``text\_0"}\\
        \textit{1, ``text\_1"} ...
        
        The edges in G are:\\
        \texttt{src,edge\_attr,dst}\\
        \textit{0, ``relation\_a'', 1}\\
        \textit{2, ``relation\_b'', 3} ...
        
        \textbf{Question:}\\
        $\{Question\}$
        
        \textbf{Constraint:}\\
        Please provide a direct answer to the question.
        
        \textbf{Answer:}
    \end{minipage}
}
\end{center}

\paragraph{Zero-shot Output Constraints.}
To ensure the LLM outputs are compatible with our automated evaluation scripts, we inject specific formatting constraints into the \texttt{Constraint} field of the prompt. These instructions suppress Chain-of-Thought (CoT) reasoning in the output generation, forcing the model to produce the final answer directly (e.g., a single integer or a ``yes/no" token).

\subsection{Dual-Path Baseline Details}
\label{sec:appendix_dual_path}

We implement the dual-path baselines by adapting the GraphToken paradigm~\cite{perozzi2024let} to incorporate explicit textual grounding. Specifically, given a graph $\mathcal{G}=(\mathcal{V}, \mathcal{E})$, structural features are encoded by a 2-layer GCN~\cite{kipf2016semi} ($d=128$). Crucially, discrete node identifiers and connectivity triplets are simultaneously provided via the text prompt following the linearization format mentioned in Appendix~\ref{sec:appendix_graph_linear}, ensuring the model has access to both symbolic information and continuous structural signals.

To generate the continuous graph tokens $Z_\mathcal{G}$, we evaluate three readout strategies:

\begin{itemize}
    \item \textbf{$\text{GraphToken}_\text{G}$ (Graph-level):} Applies global mean pooling to the node features, followed by a linear projection. This condenses the entire structure into a single summary token, yielding $Z_\mathcal{G} \in \mathbb{R}^{1 \times D}$ (where $D$ is the hidden dimension of the LM). This architecture is structurally equivalent to the design proposed in G-Retriever~\cite{he2024g}.
    
    \item \textbf{$\text{GraphToken}_\text{N}$ (Node-level):} Projects each node's representation individually via a linear layer. This produces a sequence of $|\mathcal{V}|$ tokens, resulting in $Z_\mathcal{G} \in \mathbb{R}^{|\mathcal{V}| \times D}$.
    
    \item \textbf{$\text{GraphToken}_\text{E}$ (Edge-level):} Explicitly encodes connectivity by concatenating the source and target node features for every edge ($[h_u; h_v]$) and passing them through a linear projector. This results in a sequence of tokens $Z_\mathcal{G} \in \mathbb{R}^{|\mathcal{E}| \times D}$ that provides dense relational context.
\end{itemize}

Finally, for all variants, the generated graph tokens $Z_\mathcal{G}$ are prepended to the text embeddings $Z_\text{text}$ to form the combined input sequence $[Z_\mathcal{G}; Z_\text{text}]$ for the language model.

%%%%%%%%%%%%%%%%%%%%%%%%%%%%%%%%%%%%%%%%%%%%%%%%%%%%%%%%%%%%%%%%%%%%%%%%%%%%%%%
%%%%%%%%%%%%%%%%%%%%%%%%%%%%%%%%%%%%%%%%%%%%%%%%%%%%%%%%%%%%%%%%%%%%%%%%%%%%%%%

\end{document}